\pdfoutput=1
\documentclass[11pt]{article}

\usepackage[]{naacl2021}

\usepackage{times}
\usepackage{latexsym}
\usepackage[T1]{fontenc}
\usepackage[utf8]{inputenc}
\usepackage{microtype}

\usepackage{etoolbox,siunitx}
\usepackage{graphicx}
\usepackage{subfigure}
\usepackage{booktabs} % for professional tables
\usepackage{multirow}
\usepackage{amsmath}
\usepackage{float}
\usepackage{amssymb}
\usepackage{mathtools}
\usepackage{algorithm}
\usepackage{algorithmic}
\usepackage{caption}
\usepackage{url}
\usepackage{xspace}

\DeclareMathOperator*{\argmax}{arg\,max}

\hypersetup{colorlinks=false}%hidelinks}

\newcommand{\sys}{FLIN\xspace}
\newcommand{\intdataset}{WebNav\xspace}
\newcommand{\extdataset}{DialQueries\xspace}

\newcommand{\id}[1]{{\sf\small #1}}

\newcommand{\myurl}[1]{{\footnotesize\url{#1}}}
\renewcommand{\paragraph}[1]{\vspace{0.4ex}\noindent\textbf{#1}}

\title{FLIN: A Flexible Natural Language Interface for Web Navigation}  

\author{Sahisnu Mazumder\thanks{~~Work done while interning at Microsoft Research.} \\
 Department of Computer Science  \\
 University of Illinois at Chicago, USA \\
 \texttt{sahisnumazumder@gmail.com} \\\And
 Oriana Riva \\
 Microsoft Research \\
 Redmond, USA \\
 \texttt{oriana.riva@microsoft.com} \\}

\date{}

\begin{document}
\maketitle

\begin{abstract}
AI assistants can now carry out tasks for users by directly interacting with website UIs. Current semantic parsing and slot-filling techniques cannot flexibly adapt to many different websites without being constantly re-trained. 
%However, training a natural language interface that maps a user commands to website actions is challenging for existing semantic parsing techniques due to the variable and unknown set of actions that characterize websites. 
We propose \textit{\sys}, a natural language interface for web navigation that maps user commands to concept-level actions (rather than low-level UI actions), thus being able to flexibly adapt to different websites and handle their transient nature. We frame this as a ranking problem: given a user command and a webpage, \sys learns to score the most relevant navigation instruction (involving action and parameter values). To train and evaluate \sys, we collect a dataset using nine popular websites from three domains. Our results show that \sys was able to adapt to new websites in a given domain. 
%Quantitative results show that \sys learns a general model capable of adapting to new websites. 
\end{abstract}

\section{Introduction}
AI personal assistants, such as Google Assistant, can now interact directly with the UI of websites to carry out human tasks~\cite{duplex-on-web}. Users issue commands to the assistant, and this executes them by typing, selecting items, clicking buttons, and navigating to different pages in the website. Such an approach is appealing as it can reduce the dependency on third-party APIs and expand an assistant's capabilities. This paper focuses on a key component of such systems: a natural language (NL) interface capable of mapping user commands (e.g., ``\textit{find an Italian restaurant for 7pm}") into navigation instructions that a web browser can execute. 

One way to implement such an NL interface is to map user commands directly into low-level UI actions (button clicks, text inputs, etc.). The UI elements appearing in a webpage are embedded by concatenating their DOM attributes (tag, classes, text, etc.). Then, a scoring function~\cite{pasupat2018mapping} or a neural policy~\cite{liu2018reinforcement} are trained to identify which UI element best supports a given command. Learning
at the level of UI elements is effective, but only in controlled (UI elements do not change over time~\cite{pmlr-v70-shi17a}) or restricted (single applications~\cite{Branavan09}) environments. This is not the case in the ``real'' web, where (i) websites are constantly updated, and (ii) a user may ask an assistant to execute the \textit{same} task in \textit{any} website of their choice (e.g., ordering pizza with \href{https://dominos.com}{dominos.com} or \href{pizzahut.com}{pizzahut.com}). The transient nature and diversity of the web call for an NL interface that can \textit{flexibly adapt} to environments with a \textit{variable} and \textit{unknown} set of actions, without being constantly re-trained. 

\begin{figure*}[t!]
	\centering
	\includegraphics[width=\textwidth]{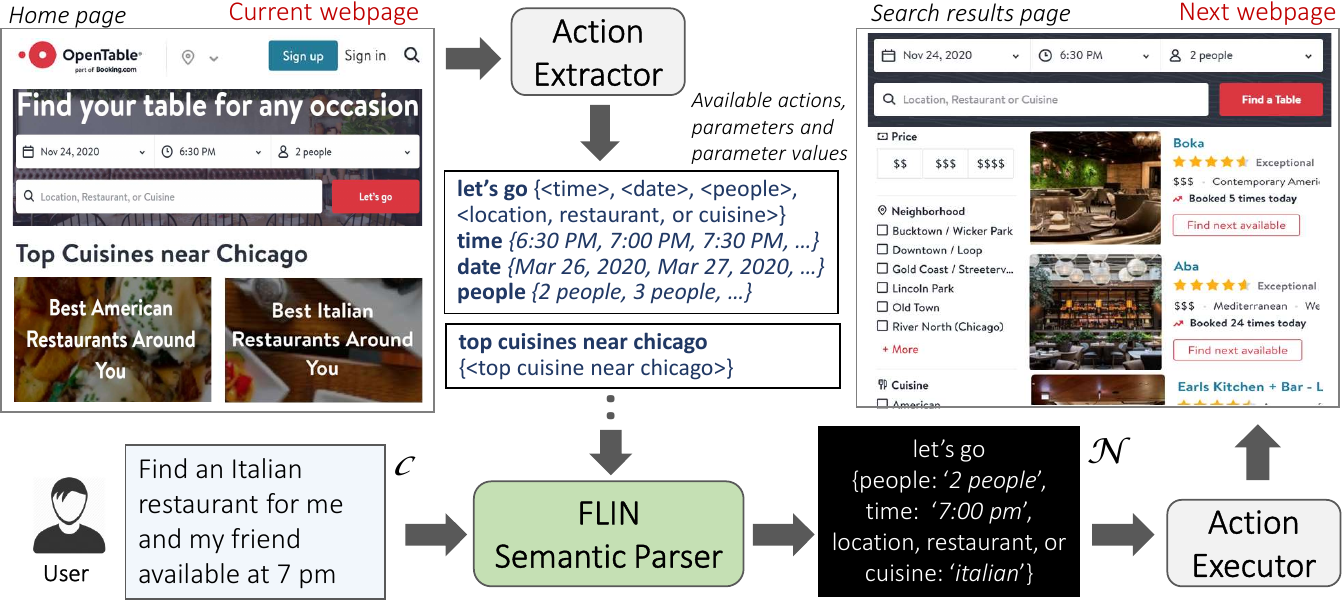}
	\caption{Web task execution driven by NL commands in the OpenTable website. The user command is mapped to the concept-level action ``Let's go'' whose execution causes the transition from the home to the search results page.}
	\label{fig:task_example}
\end{figure*}

To achieve this goal, we take two steps. First, we conceptualize a new way of designing NL interfaces for web navigation. Instead of mapping user commands into low-level UI actions, we map them into meaningful ``\textit{concept-level}" actions. Concept-level actions are meant to \textit{express what a user perceives when glancing at a website UI}. In the example shown in Figure~\ref{fig:task_example}, the homepage of \href{https://opentable.com}{OpenTable} has a concept-level action ``\textit{Let's go}" (where ``Let's go'' is the label of a search button), which represents the concept of searching something which can be specified using various parameters (a date, a time, a number of people and a search term). % (e.g., a location, a restaurant name or a cuisine type). 
Intuitively, websites in a given domain (say, all restaurant websites) share semantically-similar concept-level actions and the semantics of a human task tend to be time invariant. Hence, learning at the level of concept-level actions can lead to a more flexible NL interface. % for web navigation.

%solve the challenge of dealing with a \textit{variable} and \textit{unknown} set of actions.

However, while concept-level actions vary less than raw UI elements, they still manifest with different representations and parameter sets across websites. Searching a restaurant in \href{https://opentable.com}{opentable.com}, for example, corresponds to an action ``\textit{Let's go}'' which supports up to four parameters; in \href{https://yelp.com}{yelp.com}, the same action is called ``\textit{search}'' and supports two parameters (search term and location). Websites in one domain may also have different action types (e.g., making a table reservation vs. ordering food). % and be updated over time. 

Our second insight to tackle this problem is to leverage semantic parsing methods in a novel way. Traditional semantic parsing methods~\cite{zelle1996learning,zettlemoyer2007online,Branavan09,lau09,thummalapenta12} deal with environments that have a \textit{fixed} and \textit{known} set of actions, hence cannot be directly applied. Instead, we propose \textit{\sys}, a new semantic parsing approach where instead of learning how to map NL commands to executable logical forms (as in traditional semantic parsing), we leverage the semantics of the symbols (name of actions/parameters and parameter values) contained in the logical form (the navigation instruction) to \textit{learn how to match it} against a given command. Specifically, we model the semantic parsing task as a \textit{ranking problem}. Given an NL command $c$ and the set of actions $A$ available in the current webpage, \sys scores the actions with respect to $c$. Simultaneously, for each parameter $p$ of an action, it extracts a phrase $m$ from $c$ that expresses a value of $p$, and then scores $p$'s values with respect to $m$ to find the best value assignment for $p$. Each action with its associated list of parameter value assignments represents a candidate navigation instruction to be ranked. \sys learns a net score for each instruction based on corresponding action and parameter value assignment scores, and outputs the highest-scored instruction as the predicted navigation instruction. %We formally introduce the semantic parsing problem for web navigation in \S\ref{sec:problem} and discuss the details of \sys in \S\ref{sec:approach}. 

To collect a dataset for training and testing \sys, we built a simple rule-based \textit{Action Extractor} tool that extracts concept-level actions along with their parameters (names and values, if available) from webpages. The implementation and evaluation of this tool is out of scope for this paper.\footnote{The tool processes a webpage's DOM tree (structure and attributes) and visual appearance (using computer vision techniques) to extract actions along with names and values of their parameters (if any). It is an active area of research~\cite{remaui15,design-semantics-uist18,chen2020object,chen20:gui-accessibility}.} In a complete system, illustrated in Figure~\ref{fig:task_example}, we envision the Action Extractor to extract and pass the concept-level actions present in the current webpage to \sys, which computes a candidate navigation instruction $\mathcal{N}$ to be executed by an \textit{Action Executor} (e.g., a web automation tool such as \citet{selenium-driver} or Ringer~\cite{ringer16}).

%Given the notion of concept-level actions, a complete system for web task execution would consist of three components, as illustrated in Figure~\ref{fig:task_example}: \textit{(1)} an \textit{Action Extractor} which extracts concept-level actions along with their parameters from webpages; \textit{(2)} a \textit{Semantic Parser} which maps user commands into navigation instructions; and \textit{(3)} an \textit{Action Executor} that executes the navigation instructions by performing low-level UI actions. In this paper, we focus on building the second component (the Semantic Parser) and assume that other two components have already been built (hence we do not evaluate them in this paper).\footnote{In our implementation, we built a rule-based Action Extractor that mines a webpage's DOM tree (including tree structure, DOM attributes and visible texts, obtained using Optical Character Recognition (OCR)) to enumerate its actions along with the names and values of their parameters (if any). We built an Action Executor using existing UI automation tools (\citet{selenium-driver} and Ringer~\cite{ringer16}).} 

%These two components incorporate website-specific knowledge, but, once built, they can be maintained semi-automatically through periodic crawling.

Overall, we make the following contributions: \textbf{(1)} we conceptualize a new design approach for NL interfaces for web navigation based on concept-level actions; \textbf{(2)} we build a match-based semantic parser to map NL commands to navigation instructions; and \textbf{(3)} we collect a new dataset based on nine websites (from restaurant, hotel and shopping domains) and provide empirical results that verify the generalizability of our approach. Code and dataset are available at~ \href{https://github.com/microsoft/flin-nl2web}{https://github.com/microsoft/flin-nl2web}.

\section{Related Work}
\label{rel-work}

Semantic parsing has long been studied in NLP with applications to databases~\cite{zelle1996learning,zettlemoyer2007online,ferre2017sparklis}, knowledge-based question answering \cite{berant2013semantic,yih2015semantic}, data exploration and visual analysis \cite{setlur2016eviza,utama2018end,lawrence2016nlmaps,gao2015datatone}, 
~robot navigation~\cite{artzi2013weakly,tellex2011understanding,janner2018representation,guu2017language,fried2017unified,garcia2018explain}, object manipulation~\cite{frank2012predicting}, object selection in commands~\cite{golland2010game,smith2013learning}, language game learning~\cite{wang2016learning}, and UI automation~\cite{Branavan09,fazzini18,zhao19-icse}. This work assumes environments with a fixed and known set of actions, while we deal with \textit{variable} and \textit{unknown} sets of actions.

Work on NL-guided web task execution includes learning from demonstrations~\cite{allen2007plow}, building reinforcement learning agents~\cite{shi2017world,liu2018reinforcement}, training sequence to sequence models to map natural language commands into web APIs \cite{su2017building,su2018natural}, and generating task flows from APIs~\cite{williams2019automatic}. These techniques assume different problem settings (e.g., reward functions) or deal with low-level web actions or API calls. Unlike \sys, they do not generalize across websites.

\citet{pasupat2018mapping} propose an embedding-based matching model to map natural language commands to low-level UI actions such as hyperlinks, buttons, menus, etc. Unlike \sys, this work does not deal with predicting parameter values (i.e., actions are un-parametrized). %In \S\ref{sec:ev-results}, the \sys-embed variant (which \sys outperforms) uses an adaption of their matching approach extended to extract parameters.

%cut: asru-XuS13
Models that jointly perform intent detection and slot filling~\cite{guo14_slotfill,liu-lane-2016-joint,Chen2019BERTFJ} are not applicable to our problem for three reasons. First, they are trained on a per-application basis using application-specific intent and slot labels, and thus cannot generalize across websites. Second, they semantically label words in an utterance, but do not do value assignment, hence they cannot output executable navigation paths. Third, they perform multi-class classification (i.e., they assume only one intent to be true for a user query) and have no notion of state (e.g., current webpage). They do not deal with intents with overlapping semantics which may occur across pages of the same website (e.g., in the example shown in Figure~\ref{fig:task_example}, the same user query may map to the action ``\textit{Let's go}" in the OpenTable's home page or to the action ``\textit{Find a Table}" in the search results page).

%However, this work deals only with un-parametrized actions, i.e., proposed a embedding based matching model to match NL commands to actions which are web page elements like  \textit{links}, \textit{buttons}, and \textit{form inputs}) and does not bother about recognizing action parameters and assigning values to those parameters (like ours). 
%We also adapt their embedding based matching approach (best performing method in their work) and build a FLIN-variant- FLIN-embed to fit the method into our problem setting. Experimental results show our proposed model FLIN achieves superior performance than FLIN-embed. 

\section{Problem Formulation}
\label{sec:problem}

Let $A_{w}=~$\{$a_1, a_2, ..., a_n$\} be the set of concept-level actions available in a webpage $w$. Each action $a$ $\in$ $A_{w}$ is defined by an \textit{action name} $n_a$ and a set of $K$ \textit{parameters} $P_a=~$\{$p_1$, $p_2$, ..., $p_K$\}. Each parameter $p$ $\in$ $P_a$ is defined by a name\footnote{We use the same notation $p$ to refer to the parameter and the parameter name interchangeably.} and a domain $dom(p)$ (i.e., a set of values that can be assigned to parameter $p$), and can be either \textit{closed domain} or \textit{open domain}. 

For closed-domain parameters, the domain is bounded and consists of a finite set of values that $p$ can take; the set is imposed by the website UI, such as the available colors and sizes for a product item or the available reservation times for a restaurant. 

For open-domain parameters, the domain is, in principle, unbounded, but, in practice, it consists of all words/phrases which can be extracted from an NL command $c$. With reference to Figure~\ref{fig:task_example}, the ``\textit{let's go}" (search) action has $n_a=$``\textit{let's go}" and $P_a =$  \{``\textit{time}", ``\textit{date}", ``\textit{people}", ``\textit{location, restaurant, or cuisine}"\}. The first three parameters are closed domain and the last one (the search term) is open domain. The Action Extractor module (Figure~\ref{fig:task_example}) names actions and parameters after labels and texts appearing in the UI (or, if absent, using DOM attributes); it also automatically scrapes values of closed-domain parameters (from drop-down menus or select lists).

Given the above setting, our goal is to map an NL command $c$ issued in $w$ into a navigation instruction $\mathcal{N}$, consisting of a correct \textit{action name} $n_{a^*}$ corresponding to action $a^*$$\in$ $ A_{w}$ and an associated list of $m$$\leq$$|P_{a^*}|$ correct \textit{parameter-value assignments}, given by \{($p_i=v'_j$) $~|~ p_i$ $\in$ $P_{a^*}$, $v'_j$ $\in$ $dom(p_i)$, $0$ $\leq$ $i$ $\leq$ $K$, $1$ $\leq j \leq |dom(p_i)|$ \}. 

%oriana: moved above
%Note that if $p_i$ is a closed-domain parameter, $v'_j$ is chosen from $dom(p_i)$, and if $p_i$ is an open-domain parameter, $v'_j$ is a word or phrase extracted from $c$ (in which case, $dom(p_i)$ consists of all possible phrases in $c$). 

\begin{figure*}[t!]
	\centering
	\includegraphics[width=\textwidth]{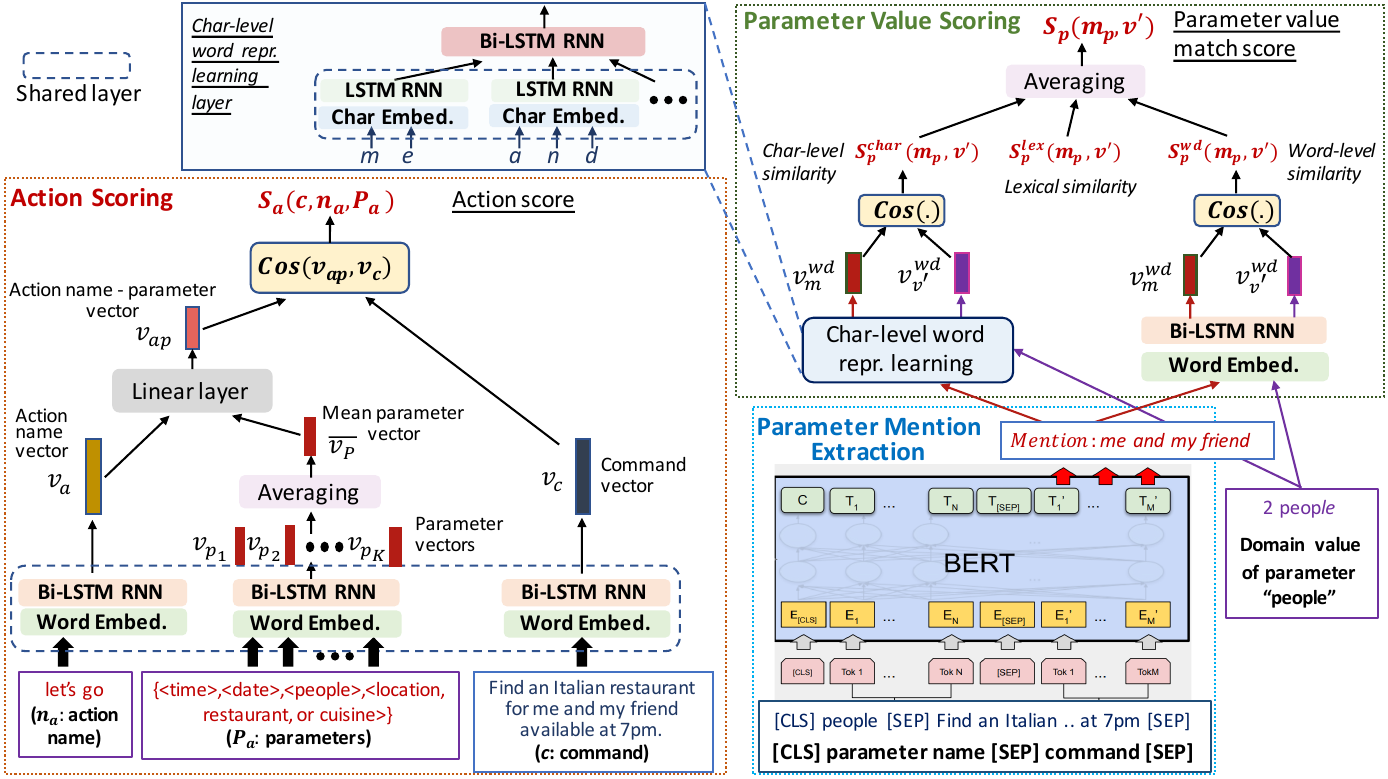}
	\caption{Architecture of \sys including Action Scoring, Parameter Mention Extraction and Parameter Value Scoring components [BERT block diagram courtesy: \cite{devlin2018bert}]. }
	\label{fig:arch}
\end{figure*}

\section{The FLIN Model}
\label{sec:approach}

The task of solving the above semantic parsing problem can be decomposed into two sub-tasks: \textbf{(i) action recognition}, i.e., recognizing the action $a \in A_{w}$ intended by $c$, and \textbf{(ii) parameter recognition and value assignment}, i.e., deciding whether a parameter of an action is expressed in $c$ and, if so, assigning the value to that parameter. A parameter \textit{is expressed} in $c$ by a mention (word or phrase). For example, in Figure~\ref{fig:task_example}, ``\textit{me and my friend}" is a mention of parameter ``\textit{people}" in $c$ and a correct parsing should map it to the domain value ``\textit{2 people}". Thus, the second sub-task involves first extracting a mention of a given parameter from $c$, and then matching it against a set of domain values to find the correct value assignment. For an open-domain parameter, the extracted mention becomes the value of the parameter and no matching is needed, e.g. in Figure~\ref{fig:task_example}, the mention ``\textit{Italian}'' should be assigned to the parameter ``\textit{location, restaurant, or cuisine}".

With reference to Figure~\ref{fig:arch}, \sys consists of four components, designed to solve the aforementioned two sub-tasks: \textbf{(1) Action Scoring}, which scores each available action with respect to the given command (\S\ref{sec:action-scoring}); \textbf{(2) Parameter Mention Extraction}, which extracts the mention (phrase) from the command for a given parameter (\S\ref{sec:par-extract}); \textbf{(3) Parameter Value Scoring}, which scores a given mention with a closed-domain parameter value or rejects it if no domain values can be mapped to the mention (\S\ref{sec:par-value}); and \textbf{(4) Inference} (not shown in the figure), which uses the scores of actions and parameter values to infer the action-parameter-value assignment with the highest score as the predicted navigation instruction (\S\ref{sec:inference}).

\subsection{Action Scoring}
\label{sec:action-scoring}

Given a command $c$, we score each action $a \in A_{w}$ to measure the similarity of $a$ and $c$'s intent. We loop over the actions in $A_{w}$ and their parameters to obtain a list of action name and parameters pairs ($n_a$, $P_a$), and then score them with respect to $c$.

To score each pair ($n_a$, $P_a$), we learn a neural network based scoring function $S_a(.)$ that computes its similarity with $c$. We represent $c$ as a sequence \{$w_1, w_2, ..., w_R$\} of $R$ words. To learn a vector representation of $c$, we first convert each $w_i$ into corresponding one-hot vectors $x_i$, and then learn embeddings of each word using an embedding matrix $E_w \in \mathbb{R}^{d \times |V|}$ as $\mathbf{v}_i$$=$$E_w.x_i$, where $V$ is the word vocabulary. Next, given the word embedding vectors \{$\mathbf{v}_i ~|~ 1 \leq i \leq |R| $\}, we learn the forward and backward representation using a Bi-LSTM network \cite{schuster1997bidirectional}. %, as:
%Bi-directional Long Short Term Memory
% \vspace{-0.1cm}
% \begin{equation}
% \label{eq1}
% 	\scalebox{0.95}{$
% 	\begin{split}
% 		\overrightarrow{\mathbf{h}_r} = \overrightarrow{LSTM}~(\mathbf{v}_i,~ \overrightarrow{\mathbf{h}_{r-1}})\\
% 		\overleftarrow{\mathbf{h}_r} = \overleftarrow{LSTM}~(\mathbf{v}_i,~ \overleftarrow{\mathbf{h}_{r+1}})
% 	\end{split}
% 	$}
% \end{equation}
% %\vspace{-0.2cm}
Let the final hidden state for forward LSTM and backward LSTM, after consuming \{$\mathbf{v}_i ~|~ 1 \leq i \leq R $\}, be $\overrightarrow{\mathbf{h}_R}$ and $\overleftarrow{\mathbf{h}_1}$ respectively, we learn a joint representation of $c$ as $\mathbf{v}_c = [\overrightarrow{\mathbf{h}_R} ; \overleftarrow{\mathbf{h}_1}] \in \mathbb{R}^{2d}$, where [;] denotes concatenation. %We consider $\mathbf{v}_c \in \mathbb{R}^{2d}$ to be the word sequence level vector representation of the command $c$.

Next, we learn a vector representation of ($n_a$, $P_a$). We use the same word embedding matrix $E_w$ and Bi-LSTM layer to encode the action name $n_a$ into a vector $\mathbf{v}_a$$=$$BiLSTM(n_a)$. Similarly, we encode each parameter $p \in P_a$ into a vector $\mathbf{v}_p$$=$$BiLSTM(p)$, and compute the \textit{net parameter semantics} of action $a$ as the mean of the parameter vectors ($\overline{\mathbf{v}_p}$$=$$ mean\{\mathbf{v}_p~|~p \in P_a\}$). Finally, to learn the overall semantic representation of ($n_a$, $P_a$), we concatenate $\mathbf{v}_a$ and $\overline{\mathbf{v}_p}$ and learn a combined representation using a feed-forward (FF) layer as
\begin{equation}
\label{eq2}
	\mathbf{v}_{ap} = tanh(W_a.[\mathbf{v}_a;\overline{\mathbf{v}_p}] + b_a)
\end{equation}

where $W_a \in \mathbb{R}^{4d \times 2d}$ and $b_a \in \mathbb{R}^{2d}$ are weights and biases of the FF layer, respectively.
%$\mathbf{v}_{ap} = tanh(W_a.[\mathbf{v}_a;\overline{\mathbf{v}_p}] + b_a)$, 

Given $\mathbf{v}_c$ and $\mathbf{v}_{ap}$, we compute the intent similarity between $c$ and ($n_a$, $P_a$) using cosine similarity: 
\begin{equation}
\begin{split}
\label{eq3}
    S_a(c, n_a, P_a) &= \frac{1}{2}~ [~cosine(\mathbf{v}_c,\mathbf{v}_{ap}) + 1]\\
    &= \frac{1}{2}~(\frac{\mathbf{v}_c.\mathbf{v}_{ap}}{||\mathbf{v}_c||||\mathbf{v}_{ap}||} + 1 )
\end{split}
\end{equation}

where $||\cdot||$ denotes euclidean norm of a vector. $S_a(\cdot) \in [0, 1]$ is computed for each $a \in A_{w}$ (and is used in inference, \S\ref{sec:inference}).

The parameters of $S_a(.)$ are learned by minimizing a margin-based ranking objective $\mathcal{L}_{a}$, which encourages the scores of each positive ($n_a$, $P_a$) pair to be higher than those of negative pairs in $w$:
\begin{equation}
\label{eq4}
	%\small
	\scalebox{0.92}{$
	\begin{split}
		\mathcal{L}_{a} = \sum_{q \in Q^+} \sum_{q' \in Q^{-}} max\{S(q') - S(q) + 0.5,~ 0\}
	\end{split}$}
\end{equation}

where $Q^+$ is a set of positive ($n_a$, $P_a$) pairs in $w$ and $Q^{-}$ is a set of negative ($n_a$, $P_a$) pairs obtained by randomly sampling action name and parameter pairs (not in $Q^+$) in $w$. %\todo{can you use a name different from $N_1$? it seems odd. I also don't understand why this is set to 1.}

\subsection{Parameter Mention Extraction} 
\label{sec:par-extract}

Given a command $c$ and a parameter $p$, the goal of this step is to extract the correct mention $m_p$ of $p$ from $c$. In particular, we aim to predict the text span in $c$ that represents $m_p$. We formulate this task as a question-answering problem, where we treat $p$ as a question, $c$ as a paragraph, and $m_p$ as the answer. We fine-tune a pre-trained BERT~\cite{devlin2018bert} model\footnote{\url{https://tfhub.dev/tensorflow/bert_en_uncased_L-12_H-768_A-12/1}} to solve this problem.

As shown in Figure~\ref{fig:arch} (bottom-right), we represent $p$ and $c$ as a pair of sentences packed together into a single input sequence of the form [CLS] $p$ [SEP] $c$, where [CLS], [SEP] are special BERT tokens. For tokenization, we use the standard WordPiece Tokenizer~\cite{wu2016google}. From BERT, we obtain $T_i$ as output token embedding for each token $i$ in the packed sequence. We only introduce a mention start vector $S \in \mathbb{R}^H$ and a mention end vector $E \in \mathbb{R}^H$ during fine-tuning. The probability of word $i$ being the start of the mention is computed as a dot product between $T_i$ %(the hidden vector for the $i^{th}$ input token) 
and $S$ followed by a softmax over all of the words in $c$: $P_i = \frac{e^{S{\cdot}T_i}}{\sum_j e^{S{\cdot}T_j}}$. The analogous formula is used to compute the end probability $P_j$. The position $i$ (position $j >$  $i$) with highest start (end) probability $P_i$ ($P_j$) is predicted as start (end) index of the mention and the corresponding tokens in $c$ are combined into a word sequence to extract $m_p$. We fine-tune BERT by minimizing the training objective as the sum of the log-likelihoods of the correct start and end positions.
We train BERT to output [CLS] as $m_p$, if $p$ is not expressed in $c$ (no mention is identified, hence $p$ is discarded from being predicted).

\subsection{Parameter Value Scoring} 
\label{sec:par-value}

Once the mention $m_p$ is extracted for a closed-domain parameter $p$, we learn a neural network based scoring function $S_p(\cdot)$ to score each $p$'s value $v'\in dom(p)$ with respect to $m_p$. If $p$ is open-domain, parameter value scoring is not needed. 

The process is similar to that of action scoring, but, in addition to \textit{word-level semantic similarity}, we also compute \textit{character-level} and \textit{lexical-level} similarity between $v'$ and $m_p$. In fact, $v'$ and $m_p$ often have partial lexical matching. For example, given the domain value ``7:00 PM" for the parameter ``\textit{time}", possible mentions may be ``\textit{\textbf{7} in the evening}'', ``\textit{19:\textbf{00} hrs}", ``\textit{at \textbf{7 pm}}", etc., where partial lexical-level similarity is observed. However, learning word-level and character-level semantic similarities is also important as ``PM" and ``evening" as well as ``7:00 PM" and ``19:00" are lexically distant to each other, but semantically closer.

\vspace{1mm}
\indent
\paragraph{Word-level semantic similarity.} We use the same word embedding matrix $E_w$ used in action scoring to learn the word vectors for both $m_p$ and $v'$. We use a Bi-LSTM layer (not shared with Action Scoring) to encode mention (value) into a word-level representation vectors $\mathbf{v}_m^{wd}$ ($\mathbf{v}_{v'}^{wd}$). We compute the word-level similarity between $m_p$ and $v'$ as
\begin{equation}
\label{eq5}
\scalebox{0.98}{$
	S_p^{wd}(m_p, v') = \frac{1}{2} ~ [~cosine(\mathbf{v}_m^{wd},\mathbf{v}_{v'}^{wd}) + 1 ]
	$}
\end{equation}
%$S_p^{wd}(m_p, v') = \frac{1}{2} ~ [cosine(\mathbf{v}_m^{wd},\mathbf{v}_{v'}^{wd}) + 1]$.

%To compute character-level similarity between $m_p$ and $v'$
%Long Short Term Memory (LSTM)

\vspace{1mm}
\indent
\paragraph{Character-level semantic similarity.} We use a character embedding matrix $E_{c}$ to learn the character vectors for each character composing the words in $m_p$ and $v'$. To learn the character-level vector representation $\mathbf{v}_m^{char}$ of $m_p$, we first learn the word vector for each word in $m_p$ by composing the character vectors in sequence using an LSTM network~\cite{hochreiter1997long}, and then compose the word vectors for all mention words using a BiLSTM layer to obtain $\mathbf{v}_m^{char}$. Similarly, the character-level vector representation $\mathbf{v}_{v'}^{char}$ of $v'$ is obtained. Next, we compute the character-level similarity between $m_p$ and $v'$ as
\begin{equation}
\label{eq6}
\scalebox{0.95}{$
	S_p^{char}(m_p, v') = \frac{1}{2} ~ [ cosine(\mathbf{v}_m^{char},\mathbf{v}_{v'}^{char}) + 1 ]
	$}
\end{equation}

%To compute lexical-level similarity between $m_p$ and $v'$

\vspace{1mm}
\indent
\paragraph{Lexical-level similarity.} We use a \textit{fuzzy string matching score} (using the Levenshtein distance to calculate the differences between sequences)\footnote{pypi.org/project/fuzzywuzzy/} and a custom \textit{value matching score} which is computed as the fraction of words in $v'$ that appear in $m_p$; then we compute a linear combination of these similarity scores (each score $\in [0, 1]$) as the net lexical-level similarity score, denoted as $S_p^{lex}(m_p, v')\in [0, 1]$.

%We learn a net similarity score between $m_p$ and $v'$. We learn it as..

\vspace{1mm}
\indent
\paragraph{Net value-mention similarity score.} It is the mean of the three scores above: $S_p(m_p, v') =mean\{S_p^{wd}(m_p, v'), S_p^{char}(m_p, v'), S_p^{lex}(m_p, v')\}$.

\vspace{1mm}
The parameters of $S_p(\cdot)$ are learned by minimizing a margin-based ranking objective $\mathcal{L}_{p}$, which encourages the scores $S_p(.)$ of each mention and positive value pair to be higher than those of mention and negative value pairs for a given $p$, and can be defined following the previously defined $\mathcal{L}_a$ (See eq.~\ref{eq4}).

\subsection{Inference} 
\label{sec:inference}

The inference module takes the outputs of Action Scoring, Parameter Mention Extraction and Parameter Value Scoring to compute a net score $S_{ap}(.)$ for each action $a \in A_{w}$ and associated list of parameter value assignment combinations. Then, it uses $S_{ap}(.)$ to predict the navigation instruction. %, as discussed below.

\vspace{1mm}
\indent
\paragraph{Parameter value assignment.} We first infer the value to be assigned to each $p\in P_a$, where the predicted value $\hat{v_p}$ for a closed-domain $p$ is given by $\hat{v_p} = \argmax_{v' \in dom(p)} S_p(m_p, v')$ \textit{provided} $S_p(m_p, v') \geq \rho$. Here, $\rho$ is a threshold score (tuned empirically) for parameter value prediction. %on a validation dataset used for training $S_p(\cdot)$. 

While performing the value assignment for $p$, we consider $S_p(m_p, \hat{v_p})$ as the confidence score for $p$'s assignment. If $S_p(m_p, v') < \rho$ for all $v' \in dom(p)$, we consider the confidence score for $p$ as $0$, implying $m_p$ refers to a value which does not exist in $dom(p)$; $p$ is discarded and no value assignment is done. If $p$ is an open-domain parameter, $m_p$ is inferred as $\hat{v_p}$ with a confidence score of $1$. 

If \textit{all} $p \in P_a$ are discarded from the prediction for a \textit{parametrized} action $a \in A_w$, we discard $a$, as $a$ no longer becomes executable in $w$.

Once we get all confidence scores for all value assignments for all $p \in P_a$, we compute the average confidence score $\overline{S_p}(P_a)$, and consider it to be the \textit{net parameter value assignment score} for $a$.

\vspace{1mm}
\indent
\paragraph{Navigation instruction prediction.} Finally, we compute the overall score for a given action $a$ and associated list of parameter value assignments as 
$S_{ap} =  \alpha * S_a(c, n_a, P_a) + (1-\alpha) * \overline{S_p}(P_a)$, where $\alpha$ is a linear combination coefficient empirically tuned. The predicted navigation instruction for command $c$ is the action and associated parameter value assignments with the highest $S_{ap}$ score.

% \vspace{-0.2cm}
% \begin{equation}
% \label{eq7}
% \scalebox{0.95}{$
% 	S_{ap} =  \alpha * S_a(c, n_a, P_a) + (1-\alpha) * \overline{S_p}(P_a)
% 	$}
% \end{equation}
% \vspace{-0.6cm}

\section{Evaluation}
\label{sec:evaluation}

We evaluate \sys on nine popular websites from three representative domains: \textit{(i) Restaurants (R)}, \textit{(ii) Hotels (H)}, and \textit{(iii) Shopping (S)}. We collect labelled datasets for each website (\S\ref{sec:dataset}) %. To evaluate the generalization capability of \sys, we 
and perform \textit{in-domain cross-website evaluation}. Specifically, we train one \sys model for each domain using one website, and test on the other (two) websites in the same domain. Ideally, a single model could be trained by using the training data of \textit{all} three domains and applied to all test websites, but we opt for \textit{domain-specific} training/evaluation to better analyze how \sys leverages the semantic overlap of concept-level actions (that exists across in-domain websites) to generalize to new websites. We discard \textit{cross-domain} evaluation because the semantics of actions and parameters do not significantly overlap across our three domains.
%\textbf{We provide a sample dataset in the Supplementary Material and will release the full dataset and code after acceptance.}

%We choose \textit{opentable.com}, \textit{hotels.com} and \textit{rei.com} for training as we observed these websites support wide-range of concept-level actions for the corresponding domains. In cross-website evaluation, we train one FLIN model for each domain, and test on other websites in the same domain. Ideally, a single FLIN model can also be trained by merging the training datasets of all three domains and can be applied to all test websites for evaluation. We choose to perform \textit{domain-specific training and evaluation} in order to better analyze- \textit{how well FLIN can leverage the semantic overlapping of concept-level actions (that exists within in-domanin websites) to achieve in-domain scalability}.

\subsection{Experimental Setup}
\label{sec:dataset}

To train and evaluate \sys, we collect two datasets: (i) \textbf{WebNav} consists of (English) command and navigation instruction pairs, and (ii) \textbf{DialQueries} consists of (English) user utterances extracted from existing dialogue datasets paired with navigation instructions. 

To collect \intdataset, given a website and a task it supports, we first identity which pages are related to the task. For example, in OpenTable we find 8 pages related to the task ``making a restaurant reservation'': the page for searching restaurants, for browsing search results, for viewing a restaurant's profile, for submitting a reservation, etc. Then, using our Action Extractor tool, we enumerate all actions present in each task-related page. For each action the extractor provides a name, parameters and parameter values, if any. The action names are inferred from various DOM attributes (\id{aria-label}, \id{value}, \id{placeholder}, etc.) and text associated with the relevant DOM element. The goal of the Action Extractor is to label UI elements as humans see them. For example, the search box in the OpenTable website, (instead of being called ``search input'') is called ``Location, Restaurant or Cuisine'', which is in fact the placeholder text associated with that input, and what users see in the UI. Parameter values are scraped automatically from DOM select elements (e.g., \id{option value} tag). We manually inspect the output of the Action Extractor and correct possible errors (e.g., missing actions). However, for every website, we obtain a different action/parameter scheme. There is no generalized mapping between similar actions/parameters across websites as building such mapping would require significant manual effort. Table~\ref{tb:dataset} reports number of pages, actions and parameters extracted for all websites used in our experiments.

With this data, we construct $<$page\_name, action\_name, [parameter\_name]$>$ triplets for all actions across all websites, and we ask two annotators to write multiple command templates corresponding to each triplet with parameter names as placeholders. A command template may be ``\textit{Book a table for $<$time$>$}''. For closed-domain parameters, the Action Extractor automatically scrapes their values from webpages (e.g., \{12:00 pm, 12:15 pm, etc.\} for the \textit{time} parameter), and we ask annotators to provide paraphrases for them (e.g., ``at noon''). For open-domain parameters, we ask annotators to provide example values (e.g., ``pizza'' for a restaurant search term). We assemble the final dataset by instantiating command templates with randomly-chosen parameter value paraphrases, and then split it into train, validation and test datasets. Overall, we generate a total of 53,520 command and navigation instruction pairs.  % The \intdataset size for all websites is in Table~\ref{tb:dataset}.
We use train and validation splits for \textit{opentable.com}, \textit{hotels.com} and \textit{rei.com} for model training. Table~\ref{tb:dataset} summarizes the sizes of the train, validation and test splits for all websites. % and test splits for these sites and other sites for in-website and cross-website evaluation. 

% R: 20,688
% H 22,324
% S: 10,508
%53,520

The second dataset, \extdataset, consists of real user queries extracted from the SGD dialogue dataset~\cite{dstc8} and from Restaurants, Hotels and Shopping ``pre-built agents'' of Dialogflow (\href{https://dialogflow.com/}{dialogflow.com}). We extract queries that are mappable to our website's tasks and adapt them by replacing out-of-vocabulary mentions of restaurants, hotels, cities, etc. with equivalent entities from our vocabulary. We manually map 421, 155 and 63 dialogue queries into navigation instructions for \href{https://www.opentable.com/}{opentable.com}, \href{https://www.hotels.com/}{hotels.com}, \href{https://www.rei.com/}{rei.com}, respectively.  We use this dataset only for evaluation purposes. 

%In this process, we manually labelled the commands to correct actions and parameter in the action graph. We also randomly select a subset of the commands and asked users to manually edited them by only replacing the parameter mentions with valid parameter mentions which can be mapped to the parameter domain values (present in the website) to adapt them for the website.  User queries that could not be mapped to any action were mapped to \id{NULL} action. Table~\ref{tb:datasets} shows the statistics of the external commands used in evaluation.

\begin{table}[t!]
	\centering
	\caption{\intdataset dataset statistics. In the last column, -/-/x denotes the website is only used for evaluation, not for training, and X is the number of commands used.}
	\label{tb:dataset}
	\scalebox{0.72}{
		\begin{tabular}{lcccc}
			\toprule
			\textbf{Website (Domain)} & \textbf{\# Pg} & \textbf{\# Act} & \textbf{\# Par} & \textbf{Train / Valid / Test} \\
			\midrule
			\href{https://www.opentable.com/}{opentable.com} (R)          &  8 &            26    &    38     &    14332 / 2865 / 1911                                  \\ 
			\href{https://www.yelp.com}{yelp.com} (R)                     &  8  &                            14        &       25                                                &    - / - / 993                                  \\
			\href{https://www.bookatable.co.uk}{bookatable.co.uk} (R)     &  7   &            14   &             19                &   - / - / 587                                   \\ 
			\hline
			\href{https://www.hotels.com}{hotels.com} (H)                 &   7   &        25    &         46          &    15693 / 3137 / 1240                              \\ %\hline
			\href{https://www.hyatt.com}{hyatt.com} (H)                   &   8    &                             17    &      48                                                 &      - / - / 1150                                \\ %\hline
			\href{https://www.radissonhotels.com}{radissonhotels.com} (H) &     7   &      20    &            42                     &  - / - / 1104                                    \\ \hline
			\href{https://www.rei.com}{rei.com} (S)                       &      11   &       25   &      40              &           7001 / 1399 / 933                        \\ %\hline
			\href{https://www.ebay.com}{ebay.com} (S)                     &   9       &    24                          &   38                                                  &    - / - / 556                                 \\ %\hline
			\href{https://www.macys.com}{macys.com} (S)                   &  11         &    26   &         37                   & - / - / 619                                     \\ \bottomrule
	\end{tabular}}
\end{table}

\begin{table*}[t!]
	\small
	\centering
	\caption{In-website and cross-website performance comparison of	\sys variants. \intdataset is used for training and/or testing, as specified. All metric scores are scaled out of 1.0.}
	\label{tb:int-tests}
%\scalebox{1}{
\begin{tabular}{l|cccc|cccc|cccc}
\toprule
           & A-acc  & P-F1  & EMA  & PA-100   & A-acc  & P-F1 & EMA    & PA-100 & A-acc & P-F1   & EMA   & PA-100   \\ 
\midrule                                          
           & \multicolumn{4}{c|}{R: opentable.com (training website)}                      & \multicolumn{4}{c|}{R: yelp.com }                           & \multicolumn{4}{c}{R: bookatable.co.uk}                  \\ 
%\hline
%ID-SF &           &           & &&&&&&&&&          \\ 
%FLIN-embed & 0.934          & 0.839          & 0.743          & 0.781          & 0.911          & 0.826          & 0.707          & 0.809          & 0.935          & \textbf{0.863}         & \textbf{0.798}          & \textbf{0.829}          \\ 
FLIN-sem   & 0.935          & 0.499          & 0.306          & 0.310          & \textbf{0.949} & 0.483          & 0.321          & 0.380          & \textbf{0.954}          & 0.415          & 0.318          & 0.339          \\ 
FLIN-lex   & 0.618          & 0.496          & 0.272          & 0.582          & 0.493          & 0.423          & 0.243          & 0.488          & 0.654          & 0.348          & 0.235          & 0.362          \\
FLIN       & \textbf{0.937} & \textbf{0.815} & \textbf{0.679} & \textbf{0.756}   & 0.926        & \textbf{0.836} & \textbf{0.703} & \textbf{0.824}   & 0.732 & \textbf{0.639} & \textbf{0.543} & \textbf{0.603} \\ 
\midrule
           & \multicolumn{4}{c|}{H: hotels.com (training website) }                        & \multicolumn{4}{c|}{H: hyatt.com }                         & \multicolumn{4}{c}{H: radissonhotels.com}                \\ 
%ID-SF &           &           & &&&&&&&&&          \\ 
%FLIN-embed & 0.879          & 0.798          & 0.567    & 0.633          & \textbf{0.944}         & \textbf{0.686}          & 0.229    & 0.385          & 0.504          & \textbf{0.354}          &  \textbf{0.171}          & \textbf{0.242}          \\ 
FLIN-sem   & 0.933          & 0.749          & 0.423          & 0.468          & \textbf{0.806}          & 0.505          & 0.154          & 0.286          & \textbf{0.883}          & \textbf{0.630}         & \textbf{0.221}          & 0.399          \\
FLIN-lex   & 0.859          & 0.720          & 0.311          & 0.811          & 0.675          & 0.467          & 0.101          & 0.536          & 0.364          & 0.250          & 0.067          & 0.325          \\ 
FLIN       & \textbf{0.939} & \textbf{0.874} & \textbf{0.643}   & \textbf{0.869}        & 0.740 & \textbf{0.551} & \textbf{0.187}   & \textbf{0.570}    & 0.455 & 0.353 & 0.146 & \textbf{0.413}  \\ 
\midrule
           & \multicolumn{4}{c|}{S: rei.com (training website)}                           & \multicolumn{4}{c|}{S: ebay.com}                          & \multicolumn{4}{c}{S:macys.com}  \\ 
%ID-SF &           &           & &&&&&&&&&          \\ 
%FLIN-embed & 0.934  &  0.871  &   0.741  &  0.843 & 0.901          & 0.702          & 0.318          & 0.390          & 0.823          & 0.517          & 0.095          & 0.318          \\ 
FLIN-sem   & 0.852          & 0.786          & \textbf{0.668}          & 0.704          & 0.559          & 0.414          & 0.239          & 0.298          & 0.833          & 0.526          & 0.119          & 0.218          \\ 
FLIN-lex   & 0.875    & 0.781  & 0.530 & \textbf{0.828}          & \textbf{0.902} & 0.721 & 0.248 & 0.437   & 0.873 & \textbf{0.604}          & 0.166          & \textbf{0.411}          \\ 
FLIN       & \textbf{0.913}          & \textbf{0.826}          & \textbf{0.668}          & 0.782          & 0.897          & \textbf{0.729}          & \textbf{0.327}          & \textbf{0.453}          & \textbf{0.878}          & 0.598 & \textbf{0.176} & 0.358 \\ 
\bottomrule
\end{tabular}
%}
\end{table*}

\vspace{1mm}
\indent
\paragraph{Training details.} All hyper-parameters are tuned on the validation set. Batch-size is 50. Number of training epochs for action scoring is 7, for parameter mention extraction is 3, and for parameter value scoring is 22. One negative example is sampled for $Q^-$ (in Eq.~\ref{eq4}) in every epoch. Dropout is 0.1. Hidden units and embedding size are 300. Learning rate is 1e-4. Regularization parameter is 0.001, $\rho=$ 0.67 and $\alpha=$ 0.4 (\S \ref{sec:inference}). The Adam optimizer~\cite{kingma2014adam} is used for optimization. We use a Tesla P100 GPU and TensorFlow for implementation.

\vspace{1mm}
\indent
\paragraph{Compared models.} There is no direct baseline for this work as related approaches differ in the type of output or problem settings. As discussed in \S\ref{rel-work}, \citet{pasupat2018mapping} do not perform parameter recognition and value assignments. \citet{liu2018reinforcement} require a reward function for neural policy learning. Joint intent detection and slot filling models perform multi-class classification and do not consider the current state (current webpage), thus not being able to deal with similar intents in different webpages; further, they perform slot filling (equivalent to parameter mention extraction) but do not perform parameter value assignments, thus being unable to output executable paths for web navigation.

Nonetheless, we compare \sys against two of its variants that use its match-based semantic parsing approach but with the following differences: %(i) \textbf{FLIN-embed} uses an embedding-based matching function for action scoring and parameter value scoring which is adapted following the embedding-based model by \citet{pasupat2018mapping} (We use cosine similarity for computing scores); 
(i) \textbf{FLIN-sem} uses only word-level and character-level semantic similarity for parameter value scoring (no lexical similarity); and %This version thus performs matching based on semantic similarity learning.
(ii) \textbf{FLIN-lex} uses only lexical similarity in parameter value scoring. % (no word-level and character-level semantic similarity). %This version thus performs matching via lexical string similarity matching. 

%\textbf{(4) FLIN:} Our proposed model described in \S\ref{sec:approach}. 

\vspace{1mm}
\indent
\paragraph{Evaluation metrics.} 
We use accuracy (\textbf{A-acc}) to evaluate action prediction and average F1 score (\textbf{P-F1}) to evaluate parameter prediction performance. P-F1 is computed using the average \textit{parameter precision} and \textit{parameter recall} over test commands\footnote{The P-F1 (parameter F1 score) is computed as the “harmonic mean” of average parameter precision and average parameter recall (averaged over all test queries).}. Given a command, parameter precision is computed as the fraction of parameters in the predicted instruction which are correct and parameter recall as the fraction of parameters in the gold instruction which are predicted correctly. If the predicted action is incorrect or no action has been predicted for a given test command, we consider both parameter precision and recall to be 0 for the command. 

We also compute (i) \textit{Exact Match Accuracy} (\textbf{EMA}), defined as the percentage of test commands where the predicted instruction exactly matches the gold navigation instruction, and (ii) 100\% Precision Accuracy (\textbf{PA-100}), defined as the percentage of test commands for which the parameter precision is 1.0 and the predicted action is correct, but parameter recall $\leq$ 1.0. %EMA evaluates how often the model is able to \textit{fully} map the command into the gold navigation instruction and PA-100 how often it generates correct (i.e., no wrong predictions), but incomplete navigation instructions.
\footnote{Although we formulate the mapping problem as a ranking one, we do not consider standard metrics such as mean average precision (MAP) or normalized discounted cumulative gain (NDCG) because \sys outputs only one navigation instruction (instead of a ranked list), given that in a real web navigation system only one predicted action can be executed.} Similar to parameter precision and recall, while computing EMA and PA-100 for test commands, if the predicted action is incorrect or no action has been predicted for a test command, we consider both the exact match and PA-100 value to be 0 for that command.

\subsection{Performance Results}
\label{sec:ev-results}

Table~\ref{tb:int-tests} reports the performance comparison of \sys on the \intdataset dataset. We evaluate both in-website (model trained and tested on the same website, 2$^{nd}$--5$^{th}$ columns) and cross-website (model trained on one website and tested on a different one, 6$^{th}$--13$^{th}$ columns) performance. \sys and its two variants adapt relatively well to previously-unseen websites thanks to \sys's match-based semantic parsing approach. \sys achieves best overall performance, and is able to adapt to new websites by achieving comparable (or higher) action accuracy (A-acc) and parameter F1 (P-F1) score. Considering PA-100, in the Restaurants domain, 75.6\% of commands in the training website (OpenTable), and 60.3\% (bookatable) and 82.4\% (yelp) of commands in the two test websites are mapped into correct and executable actions (no wrong predictions). PA-100 is generally high also for the other two domains. EMA is lower than PA-100, as it is much harder to predict \textit{all} parameter value assignments correctly.
Regarding the \sys variants, %\sys-embed only learns the embedding of words and averages the embedding vectors to learn the overall semantics. Its performance is sometimes better than \sys (in case of bookatable and radissonhotels), %but mainly for in-website evaluation. 
%however due to its simplistic semantics learning, its performance suffers in overall.
%Due to its simplistic semantics learning, it cannot adapt to new websites as well as \sys. 
both \sys-sem and \sys-lex generally perform worse than \sys because by combining both lexical and semantic similarity \sys can be more accurate in doing parameter value assignments and generalize better.

From a generalizability point of view, the most challenging domain is Hotels. While the performance of action prediction (A-acc) for Hotels is in the 45.5\%--93.9\% range, EMA is in the 14.6\%--64.3\% range. The drop is mainly due to commands with many parameters (e.g., check-in date, check-out date, number of rooms, etc.), definitely more than in the queries for the other two domains. Shopping is more challenging than Restaurants because while in the Restaurants domain, \sys must deal with actions/parameters that relate to the \textit{same} entity type (restaurant) with a relatively-contained vocabulary, shopping products can range so widely to be effectively \textit{different entity types} with a diverse set of actions/parameters that can vary significantly across websites. For a more detailed discussion of both aspects see \S \ref{sec:ev-error}.

\begin{table}[t!]
%\small
	\centering
	\caption{Performance on the \extdataset dataset (\sys models trained on the \intdataset dataset).} % Metric scores are scaled out of 1.0.}
	\label{tb:ex-tests}
	\scalebox{0.85}{
        \begin{tabular}{lcccc}
        \toprule
                 & A-acc & P-F1  & EMA   & PA-100 \\ 
        \midrule
        opentable.com (R) & 0.719 & 0.582 & 0.565 & 0.584  \\ 
        %\midrule
        hotels.com (H) & 0.730 & 0.514 & 0.381 & 0.500 \\ 
        rei.com (S)    & 0.507 & 0.464 & 0.428 & 0.460  \\
        \bottomrule
    \end{tabular}
}
\end{table}

% \begin{table}[t!]
% \small
% 	\centering
% 	\caption{Error analysis of 135 commands from \intdataset. Columns do not sum up to 100\% as parameters in the same command may fail for different reasons. (``OD'' means open-domain, ``CD'' closed-domain).}
% 	\vspace{-0.2cm}
% 	\label{tb:errors}
% 	\scalebox{0.89}{
%         \begin{tabular}{p{3.9cm}cccc}
%         \toprule
%          Error Type     & \% (R) &\% (H) & \% (S) & \% (all) \\ % & Example \\ 
%         \midrule
%         Action not predicted    & 17.7   &  8.9   & 13.3   & 13.3 \\ %\hline 
%         Action miss-predicted   & 17.7  &  28.8    &  40.0  & 28.9 \\ %\hline
%         Fail to identify CD parameters  & 20.0  &  57.7 &  31.1  & 36.3\\ 
%         %\hline
%         CD param value miss-predicted &  31.1  & 6.7   &  4.4 &14.1 \\
%         %\hline
%         Fail to extract OD param value  & 15.5 & 11.1 & 24.4  &  17.0 \\ 
%         \bottomrule
% \end{tabular}
% }
% \vspace{-0.35cm}
% \end{table}

%FLIN-sem only learns word-level and character-level semantics of parameter values. However, they do not use lexical similarity. On the other hand, FLIN-lex only leverages lexical similarity, but does not learn semantics of parameter values. Thus, 

%Overall, the cross-website performance of FLIN shows the model achieves scalability in semantic parsing of web actions. 

%oriana
% mention integrating with knowledge bases to enlarge the vocabulary
% comment on parameter values
%error analysis
%- action accuracy of radissonhotels.com

We also test \sys on the real user queries of the \extdataset dataset available for three websites. As Table~\ref{tb:ex-tests} shows, despite \sys not being trained on \extdataset, overall its A-acc is above 50\% and its PA-100 is above 46\% which demonstrate the robustness of \sys in the face of new commands. % from Diag-dataset.

%The 135 commands includes 15 commands per website, which aggregates to 45 commands for each of the three domains- R, H and S.

%Table shows the error types and \% errors for each types for each domain and over all domains.

\subsection{Error Analysis}
\label{sec:ev-error}

% We randomly sampled 135 wrongly-predicted \intdataset commands (15 for each of the 9 websites), and classified them into 5 error types (see Table~\ref{tb:errors}). 

% Overall, 13\% of the errors happened when an action was not predicted and 29\% when an action was miss predicted, mainly due to actions in the same page with overlapping semantics (e.g., ``\textit{Options for new york for 6 people and 2 kids}" got mapped to ``\textit{select hot destination}\{`destination'= `new york'\}" instead of ``\textit{find hotel}\{`adults'=`6'; `children'=`2'; `destination'=`new york'\}"). Failures in identifying close-domain (CD) parameters were most common in $H$ and $S$ websites which tend to have a more diverse action/parameter space than $R$ sites, thus leading to action/parameter types that were not observed in training (e.g., eBay has an action ``\textit{filter by style}'' not present in Rei (training site)). Failures in predicting values of correctly-identified CD parameter mentions were often due to morphological variations in parameter values rarely observed in training (e.g.,``8:00 in the evening" got mapped to ``18:00''). Extracting open-domain (OD) parameters mainly failed due to parameter names too generic (e.g. ``\textit{search keyword}"), extracted mentions partially matching gold mentions (e.g., ``\textit{hyatt}" vs. the gold mention ``\textit{hyatt regency grand cypress}"), or parameter value formats (e.g., phone numbers).

% ============================

We randomly sampled 135 wrongly-predicted \intdataset commands (15 for each of the 9 websites), and classified them into 5 error types (see Table~\ref{tb:errors}). 

Overall, 13\% of the failures were cases in which an action was not predicted (e.g., for the command ``\textit{only eight options}" with ground truth ``\textit{filter by size}\{`size'=`8'\}", no action was predicted). 29\% of the failures were action miss-predictions (the predicted action did not match the gold action) mainly caused by multiple actions in the given webpage having overlapping semantics. E.g., the command ``\textit{options for new york for just 6 people and 2 kids}" got mapped to ``\textit{select hot destination}\{`destination'= `new york'\}'' instead of ``\textit{find hotel}\{`adults'=`6'; `children'=`2'; `destination'=`new york'\}". Similarly, ``\textit{apply the kids' shoes filter}" got mapped to ``\textit{filter by gender}\{`gender'= `kids'\}'' instead of ``\textit{filter by category}\{`category': `kids footwear'\}''.

Together with action miss-predictions, failures in identifying closed-domain parameters (third row in Table~\ref{tb:errors}) were the most common, especially in hotels and shopping websites. This is because these in-domain websites tend to have a more diverse action and parameter space than that for restaurant websites, thus leading to action and parameter types that were not observed in training data. For example, the search action in Hyatt has \textit{special rates} and \textit{use points} parameters, not present in that of Hotels.com (training site); or eBay has an action \textit{filter by style} not present in Rei (training site).

\begin{table}[t!]
\small
	\centering
	\caption{Error analysis based on 135 test commands from \intdataset. Columns do not sum up to 100\% as multiple parameters in the same command may be problematic for different reasons.}
	\label{tb:errors}
	\scalebox{1}{
        \begin{tabular}{p{3cm}SSSS}
        \toprule
         Error Type     &  \%(R) &\% (H) & \% (S) & \% (all) \\ % & Example \\ 
        \midrule
        Action not predicted    & 17.7   &  8.9   & 13.3   & 13.3 \\ \hline 
        Action miss-predicted   & 17.7  &  28.8    &  40.0  & 28.9 \\ \hline
        Failed to identify closed-domain parameter  & 20.0  &  57.7 &  31.1  & 36.3\\ 
        \hline
        Closed-domain parameter value miss-predicted &  31.1  & 6.7   &  4.4 &14.1 \\
        \hline
        Fail to extract open-domain parameter value  & 15.5 & 11.1 & 24.4  &  17.0 \\ 
        \bottomrule
\end{tabular}
}
\end{table}

Failures in predicting the value of a correctly-identified closed-domain parameter mention (forth row in the table) were often due to morphological variations in parameter values not frequently observed in training (e.g. ``\textit{8:00 in the evening}" got mapped to the value `\textit{18:00}' instead of `\textit{20:00}').

Errors in extracting open-domain parameters were due to parameter names too generic (e.g. ``\textit{search keyword}"), extracted mentions partially matching gold mentions (e.g., ``\textit{hyatt}" vs. the gold mention ``\textit{hyatt regency grand cypress}"), or multiple formats of the parameter values (e.g., various formats for \textit{phone number} or \textit{zip code}).

\section{Conclusion}
To generalize to many websites, NL-guided web navigation assistants require an NL interface that can work with new website UIs without being re-trained each time. To this end, we proposed \sys, a matching-based semantic parsing approach that maps user commands to concept-level actions. While various optimizations are possible, \sys adapted well to new websites and delivered good performance. We have used it in restaurant, shopping and hotels websites, but its design can apply to more domains.

\section*{Acknowledgments}
We would like to thank Jason Kace (Microsoft) for many constructive discussions and for designing and implementing the \textit{Action Extractor} and \textit{Action Executor} modules used by the NL-guided web navigation system.

% \appendix
% \section*{Appendix: Action/parameter extraction from websites}
% {\color{blue} The action names are extracted directly from the DOM tree of each webpage. We infer the action names directly from the (OCR visible) text associated with the relevant DOM elements using simple rules and tags such as “aria-label”, “value”, “placeholder”, “alt”, etc . The goal of the extractor is to label UI elements as humans see them. For example, the search box in the OpenTable website (instead of being called “search input”) is called “Location, Restaurant or Cuisine” - which is in fact the placeholder text associated with that input which users see in the GUI. Likewise, parameters are scraped automatically from the DOM select elements (e.g., “<option value=”3”> 3 people </option>”). There is no generalized mapping between similar actions/parameters across websites. This was an important requirement for us because building such mapping would require significant manual work.}

\bibliography{custom}
\bibliographystyle{acl_natbib}

\end{document}